# Deep Residual Bidir-LSTM for Human Activity Recognition Using Wearable Sensors

Yu Zhao[a], Rennong Yang[a], Guillaume Chevalier[b], Maoguo Gong[c]

[a]Aeronautics and Astronautics Engineering College, Air Force Engineering University, Xi'an 710038, China

[b]2325 Rue de l'Universite, Laval University, Quebec G1V 0A6, Canada

[c]International Research Center for Intelligent Perception and Computation, Xidian University, Xi'an, Shaanxi Province 710071, China

**Abstract**

Human activity recognition (HAR) has become a popular topic in research because of its wide application. With the development of deep learning, new ideas have appeared to address HAR problems. Here, a deep network architecture using residual bidirectional long short-term memory (LSTM) cells is proposed. The advantages of the new network include that a bidirectional connection can concatenate the positive time direction (forward state) and the negative time direction (backward state). Second, residual connections between stacked cells act as highways for gradients, which can pass underlying information directly to the upper layer, effectively avoiding the gradient vanishing problem. Generally, the proposed network shows improvements on both the temporal (using bidirectional cells) and the spatial (residual connections stacked deeply) dimensions, aiming to enhance the recognition rate. When tested with the Opportunity data set and the public domain UCI data set, the accuracy was increased by 4.78% and 3.68%, respectively, compared with previously reported results. Finally, the confusion matrix of the public domain UCI data set was analyzed.

## 1. Introduction

In real life, many problems can be described as time series problems. Indeed, human activity recognition (HAR) is of value in both theoretical research and actual practice. It can be used widely, including in health monitoring [1][2], smart homes [3][4], and human–computer interactions [5][6]; for example, LSTM cells are a good choice for solving HAR problems. Unlike traditional algorithms, LSTM can catch relationships in data on the temporal dimension without having to mix the time steps together as a 1D convolutional neural network (CNN) would do. As more of what is commonly called "big data" emerges, LSTM architecture can offer great performance and many potential applications.

More specifically, HAR is the process of obtaining action data with sensors; it symbolizes the action information and then allows understanding and extraction of the motion characteristics, which is what activity recognition refers to. Because of the spatial complexity and temporal divergence of behavior, there is no unified recognition method. A public domain benchmark of HAR has been introduced, and different methods of recognition have been analyzed [7]. The results showed that the K-Nearest Neighbor (KNN) algorithm outperforms other algorithms in most recognition tasks. Support Vector Machine (SVM) is another outstanding algorithm. A Multi-Class Hardware-Friendly Support Vector Machine (MC-HF-SVM), which uses fixed-point arithmetic for HAR instead of the typical floating-point arithmetic, has been proposed for sensor data [8]. Unlike the manual filtering features in previous algorithms, a systematic feature learning method that combines feature extraction with CNN training has also been proposed [9]. Subsequently, DeepConvLSTM networks outperformed previous algorithms in the Opportunity Challenge by an average of 4% of the F1 score [10]; the effects of parameters on the final result were also analyzed.

Although researchers have made great strides in HAR, room for improvement remains. Inspired by previous neural networks' architectures, we describe a novel Deep Residual Bidirectional Long Short-term Memory LSTM (Deep-Res-Bidir-LSTM) network. The deep LSTM has improved learning ability and, despite the time required to reach maximum accuracy, shows good accuracy early in training; it is especially suitable for complex, large-scale HAR problems where sensor fusion would be required. Residual connections and bidirectional communication through time are available to ensure the integrity of information flowing deeply through the neural network.

In recent years, deep learning has shown applicability to many fields, such as image processing [11][12], speech recognition [13][14][15], and natural language processing [16][17]. In ILSVRC 2012, AlexNet [18], proposed by Alex Krizhevsky, took first place, and, since then, deep learning has been considered to be applicable to solving real problems and has done so with impressive accuracy. Indeed, deep learning has become a popular area for scientists and engineers.

Another event in 2016 that drew considerable attention was the century man–machine war at the end of the game in which AlphaGo achieved victory. This event also demonstrated that deep learning, based on big data, is a feasible way to solve the non-deterministic polynomial problem.

LSTM cells, which were first proposed by Juergen Schmidhuber in 1997 [19], are variants of recurrent neural networks (RNNs). They have special inner gates that allow for consistently better performance than RNN on a time series. Compared with those of other networks, such as CNN, restricted Boltzmann machines (RBM), and auto-encoder (AE), the structure of the LSTM renders it especially good at solving problems involving time series, such as those related to natural language processing, speech recognition, and weather prediction, because its design enables gradients to flow through time readily.

Section 2 presents the baseline LSTM, Bidir-LSTM, and residual networks. In Section 3, we provide an explicit introduction to the preprocessing in HAR and describe Deep-Res-Bidir-LSTM. Several experiments were performed with HAR benchmarks: the public domain UCI data set and the Opportunity data set. We compare the accuracy of recognition of our algorithm with those of other algorithm. Finally, we summarize the research and discuss our future work.

## 2. Background

### 2.1. Baseline LSTM

LSTM [18] is an extension of recurrent neural networks. Due to its special architecture, which combats the vanishing and exploding gradient problems, it is good at handling time series problems up to a certain depth.

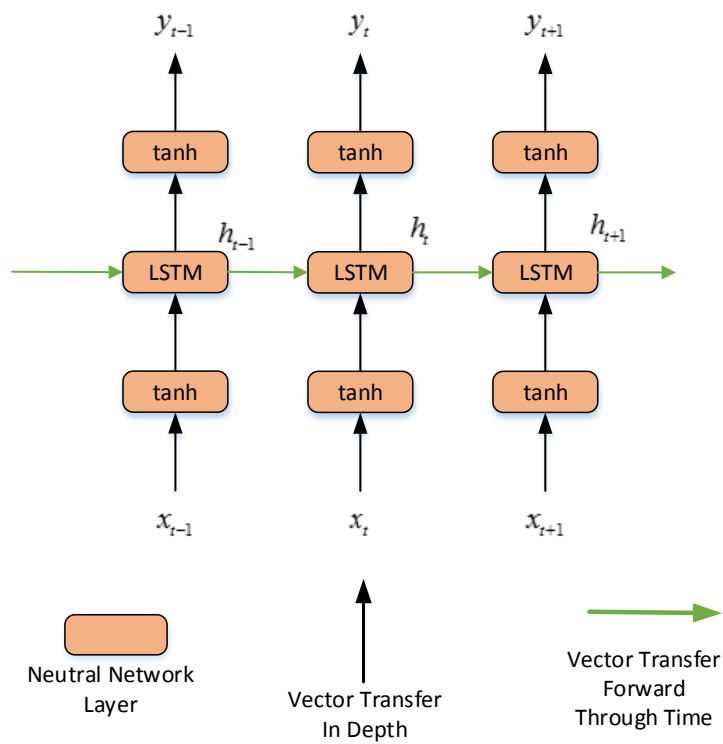

Figure 1. The unfolded structure of one-layer baseline LSTM is shown. Baseline LSTM structure operating through the time axis, from left to right.

In Figure 1, We define the input set as $\{x_0, x_1, ..., x_t, x_{t+1}, ...\}$, the output set as

$\{y_0, y_1, ..., y_t, y_{t+1}, ...\}$, and hidden layers as $\{h_0, h_1, ..., h_t, h_{t+1}, ...\}$. Then, $U, W, V$ denote weight metrics from the input layer to the hidden layer, from the hidden layer to the hidden layer, and from the hidden layer to the output layer, respectively. The transfer process of the network can be described as follows: the input tensor is transformed, along with the tensor of the hidden layer (at the last stage), to the hidden layer by a matrix transformation. Then, the output of the hidden layer passes through an activation function to the final value of the output layer.

Formally, outputs of the hidden layer and output layer can be defined as follows:

$$h_i = \begin{cases} g(\boldsymbol{U}x_i + b_i^h) & i = 0 \\ g(\boldsymbol{U}x_i + \boldsymbol{W}h_{i-1} + b_i^h) & i = 1, 2, ... \end{cases}$$
$$y_i = g(\boldsymbol{V}h_i + b_i^y) \quad i = 0, 1, ... \quad (1)$$

In theory, RNN can estimate the output of current time based on past information. However, Bengio [20] found that RNN could remember the information for only a short time, because of the vanishing and exploding gradient problems. When back propagation with a deep network is used, gradients will vanish rapidly if preventative measures that permit gradients to flow deeply are not taken. Compared with the simple input concatenation and activation used in RNNs, LSTM has a particular structure for remembering information for a longer time as an input gate and a forget gate control how to overwrite the information by comparing the inner memory with the new information arriving; this enables gradients to flow through time easily.

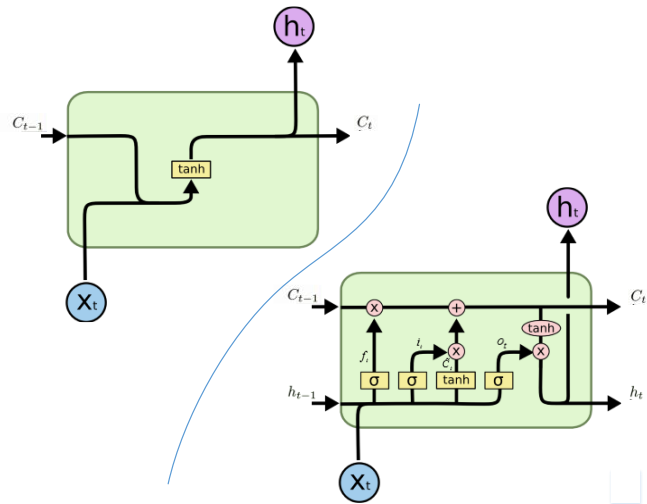

Figure 2. Comparison between a RNN and a LSTM cell. $\sigma$ represents the sigmoid function, and *tanh* represents the tanh function.

As shown in Figure 2, the input gate, the forget gate, and the output gate of LSTM are designed to control what information should be forgotten, remembered, and updated. Gating is a method to selectively pass the information that is needed. It consists of a sigmoid function and an element-wise multiplication function. The output value is between [0, 1] to allow the multiplication to then happen to let information flow or not; thus, it is considered good practice to initialize these gates to a value of 1, or close to 1, so as to not impair training at the beginning.

In the LSTM cell, each parameter at moment $t$ can be defined as follows:

$$f_t = \sigma(W_f[h_{t-1}, x_t] + b_f)$$
$$i_t = \sigma(W_i[h_{t-1}, x_t] + b_i)$$
$$\tilde{C}_t = tanh(W_c[h_{t-1}, x_t] + b_C)$$
$$C_t = f_t C_{t-1} + i_t \tilde{C}_t$$
$$o_t = \sigma(W_o[h_{t-1}, x_t] + b_o)$$
$$h_t = o_t tanh(C_t)$$

(2)

First, there is a need to forget old information, which involves the forget gate. The next step is to determine what new information needs to be kept in memory; this is done with an input gate. From that, it is possible to update the old cell state, $C_{t-1}$, to the new cell state, $C_t$. Finally, it should be decided which information should be output to the layer above with an output gate.

## 2.2. Bidirectional LSTM

In real life, human trajectories are continuous. Baseline LSTM cells can predict the current status based only on former information. It is clear that some important information may not be captured properly by the cell if it runs in only one direction.

The improvement in bidirectional LSTM is that the current output is not only related to previous information but also to subsequent information. For example, it is appropriate to predict a missing word based on context.

Bidirectional LSTM [32] is made up of two LSTM cells, and the output is determined by the two together.

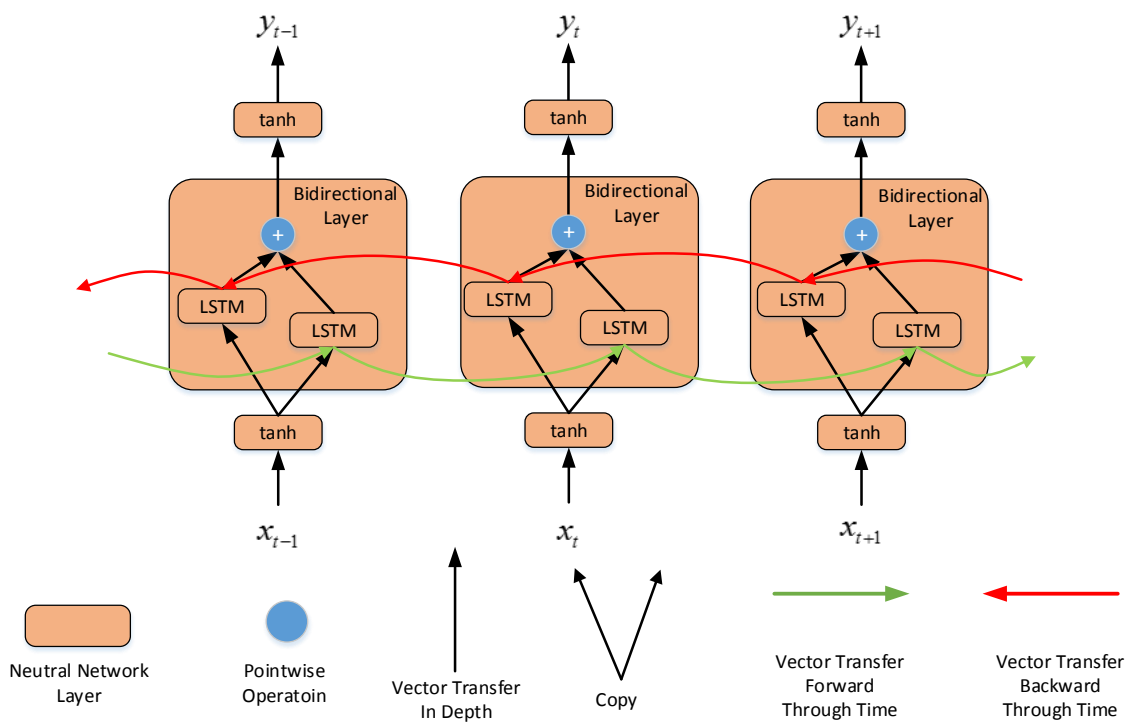

Figure 3. Standard bidirectional LSTM structure. For a bidirectional layer, it gets information from vertical direction (lower layer) and horizontal direction (past and future), and finally outputs the processed information for the upper layer.

In Figure 3, there are forward sequences $\vec{h}$ and backward sequences $\overleftarrow{h}$ in the hidden layer. For the moment, $t(t = 0, 1, 2...)$, the hidden layer and the input layer can be defined as follows:

$$\begin{aligned} \vec{h}_t &= g(\boldsymbol{U}_{\vec{h}} x_t + \boldsymbol{W}_{\vec{h}} \vec{h}_{t-1} + b_{\vec{h}}) \\ \overleftarrow{h}_t &= g(\boldsymbol{U}_{\overleftarrow{h}} x_t + \boldsymbol{W}_{\overleftarrow{h}} \overleftarrow{h}_{t-1} + b_{\overleftarrow{h}}) \\ y_t &= g(\boldsymbol{V}_{\vec{h}} \vec{h}_t + \boldsymbol{V}_{\overleftarrow{h}} \overleftarrow{h}_t + b_y) \end{aligned} \qquad (3)$$

Our bidirectional LSTM cell differs slightly from this. We concatenated the results of the two $h_t$ to then reduce the number of features in half with a ReLU fully connected hidden layer, as follows:

$$y_t = ReLU(W * concat(\vec{h}_t, \overleftarrow{h}_t) + b), \qquad (4)$$

where $concat(\cdot)$ means concatenating sequences.

**2.3. Residual Network**

The MSRA team built a 152-layer network, which was about eight times that of the VGG network [21]. Due to its excellent performance, they took first place in the 2015 ILSVRC competition owing to an absolute advantage in image classification, image location, and image detection.

As the network deepens, the research emphasis shifts to how to overcome the obstruction of information and gradient transmission. The MSRA uses residual networks. The main idea is that it is easier to optimize the residual mapping than to optimize the original, unreferenced mapping.

An important advantage of residual networks is that they are much easier to train because the gradients can be passed through the layers more directly with the addition operator that enables them to bypass some layers that would have otherwise been restrictive. This enables both better training and a deeper network, because residual connections do not impede gradients and still contribute to refining the output of a highway layer composed of such residual connections. On top of a collection of residual connections is a bottleneck where the next layers stop being residual and where a batch normalization is generally applied to normalize and restrict the usage of the feature space represented by the layer [22].

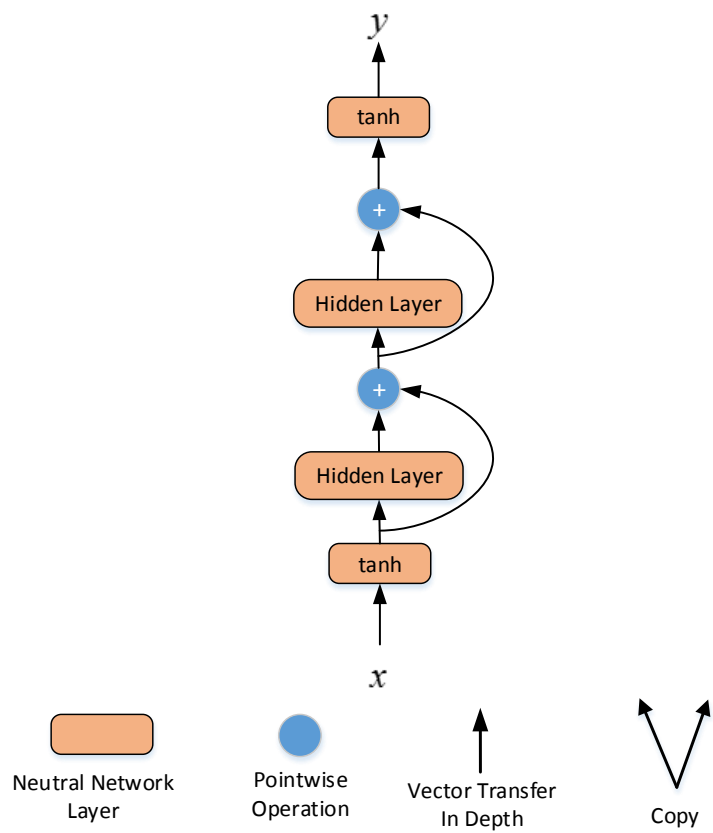

Figure 4. Isolated residual neural network

A residual network is shown in Figure 4. The lower information can transmit to upper layer directly through a highway, which increases the freedom of the information flowing. The highway structure containing skip connections can connect many supplementary $n(n=0,1,2,...)$ layers in height before the bottleneck. When $n$ equals 0, there is no residual connection: it becomes like the baseline deep-stacked LSTMs layers. In Figure 4, $n$ is 0, and the output of the hidden layer $i(i=1,2,...,L)$ can be defined as follows:

$$\begin{aligned} \boldsymbol{h}_1 &= \sigma(W_1\boldsymbol{x}+b_1) & i=1 \\ \boldsymbol{h}_i &= \sigma(W_i\boldsymbol{h}_{i-1}+b_i)+\boldsymbol{h}_{i-2} & i=2,3,...,L\text{-}1 \\ \boldsymbol{y} &= \sigma(W_y\boldsymbol{h}_i+b_y)+\boldsymbol{h}_{i-1} & i=L \end{aligned} \quad (5)$$

In the code implementation, indexing in the configuration file starts at 1 rather than 0 because we included the count of the first layer that acts as a basis before the residual cells. The same counting rule applies for indicating how many blocks of residual highway layers are stacked one on top of the other.

## 3. Our Model: Deep Residual Bidir-LSTM Network

### 3.1. Process Pipeline for HAR

The process pipeline of HAR can be divided into three parts: preprocess, training, and testing. In our case, testing was modified in parallel with training. First, testers performed activities of daily living with wearable sensors and gathered information to form the raw data set. When data were missing, we added them and then normalized to a mean of zero and standard deviation of 0.5; we then reshaped the data to fit the designed network, with windows of 128 time steps. The data were split into training and testing data sets.

Second, a training data tensor was added to the designed network so it could output a prediction. The difference between the predicted value and the real value was then compared with a sigmoid cross entropy loss with L2 weight decay to then back-propagate errors backward into the network layer by layer with the Adam Optimizer [31]. Thus, we could adjust the hyper-parameters in networks, such as the learning rate and L2 weight decay, to reduce the difference.

Finally, during testing, we added the testing tensor to the neural network architecture without affecting the learned parameters, so as to not corrupt the test. Testing did not affect the training and did not change the results. Predictions obtained from the neural network were compared with the real values. The metrics of accuracy and of the F1 score of HAR were then calculated throughout learning and, at the end, by running the tests frequently. Both the best in-training metrics and the final metrics obtained were kept for comparison.

## 3.2. Architecture of Deep Res-Bidir-LSTM Network

Considering the networks in Section 2, we proposed the Deep-Res-Bidir-LSTM to deal with HAR. Although residual connections for CNN have been used [21], this method is also available for LSTM.

Similar to building blocks, we can select modules and combine them to build a network based on our mission. The input of HAR should be a time series, and the basic structure of the LSTM guarantees that it can preserve the characteristics on the temporal dimension.

Additionally, a large network can be optimized correctly for a problem with sufficient regularization, such as L2 weight decay and dropout; however, if no regularization is used, this results in overfitting and bad operations on the test set. Complexity is good but only if countered with regularization. Too many layers and cells per layer will increase the computational complexity and waste computational resources. When the layer number and cell number reach a certain scale, the recognition accuracy will remain at a certain scale instead of increasing; by adding more depth, regularization is then needed to avoid overfitting while still improving accuracy.

Our deep LSTM neural network is limited in terms of how many data points it can access: it has access to only 128 time steps when making its predictions. Especially when deepened, the next forward/backward duo will see output from the other pass "in advance," because, logically, a backward pass for our bidirectional LSTM reverses the input and the output before the concatenation. Thus, the Bidir-LSTM has the same input and output shape as the baseline LSTM but, through depth, at a given time step, it has access to more information in advance because of the backward passes.

In general, gradient vanishing is a widespread problem for deep networks. The residual, bidirectional, and stacked layers (hence, the name "Deep Residual Bidir-LSTM") help counter this problem, because some bottom layers would otherwise be too hard to optimize when using back propagation. Combined with batch normalization on the top of each highway layer, the residual connection act as a highway for gradients; this prevents restrictions in the hidden layer feature space from being too complex and avoids outlier values at test time, combatting overfitting.

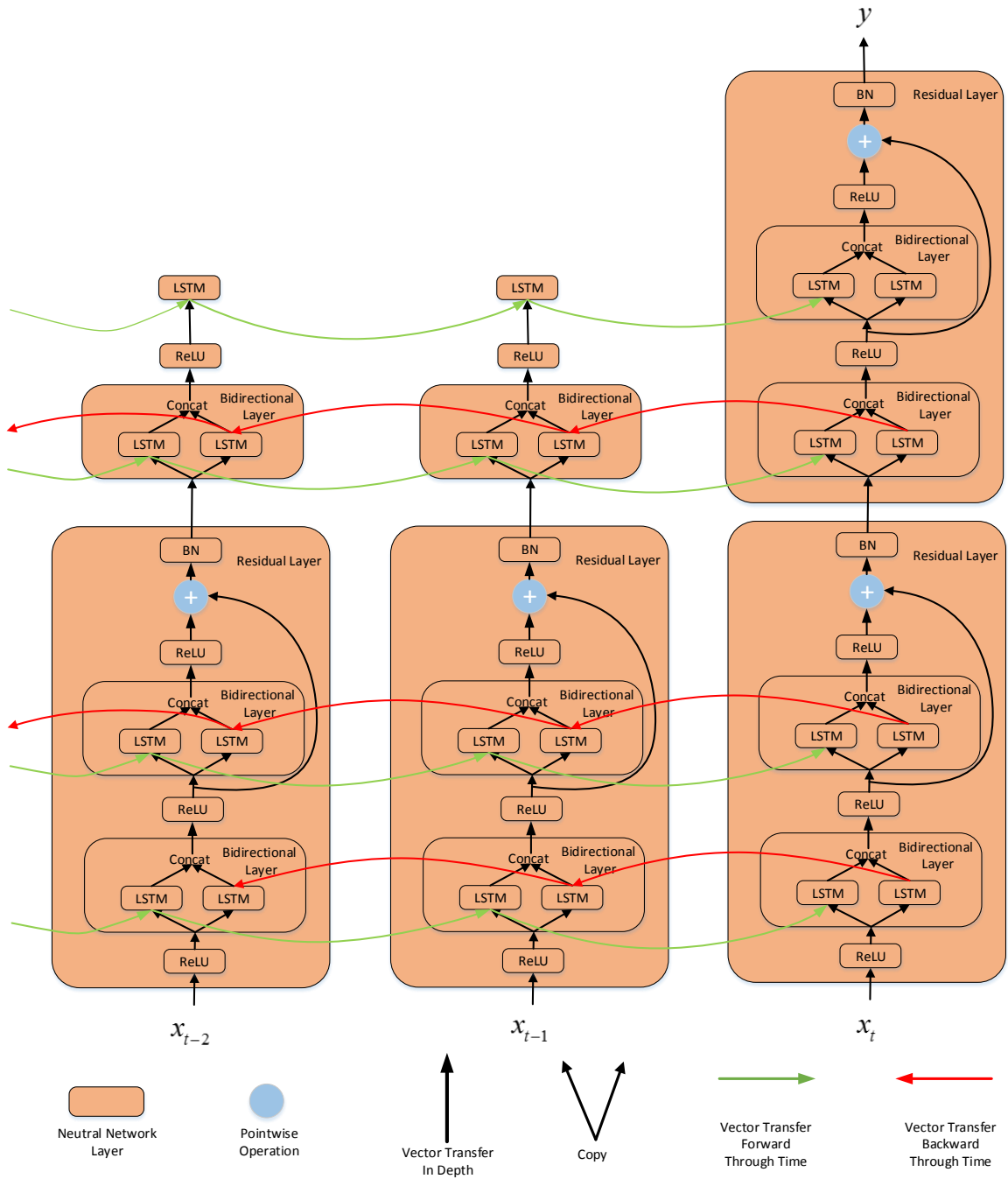

Figure 5. Unfold Deep-Res-Bidir-LSTM network architecture. "ReLU" represents the full connection layer and a ReLU function followed. And "BN" is short for batch normalization.

In Figure 5, the information flows in the horizontal direction (temporal dimension) and in the vertical direction (depth dimension). With the exception of the input and output layers, there are 2 hidden layers who has residual connection inside (hence, called "residual layer"). Moreover, each residual layer contains 2 bidirectional layers. The network in Figure 5 has a $2\times 2$ architecture, which can also be thought of as 8 LSTM cells in sum. In our network, the activity function is unified with ReLU, because it always outperforms *tanh* with deep networks to counter gradient vanishing. Although the output is a tensor for a given time window, $T$, the time axis has been crunched by the neural network. That is, we need only the last element of the output and can discard the others. Thus, only the gradient from the prediction at the last time step is applied. This also causes a LSTM cell to be unnecessary: the uppermost backward LSTM in the bidirectional pass. Hopefully, this is not of great concern because TensorFlow, the library we use, should evaluate what to compute and what not to compute. Additionally, the training data set should be shuffled during the training process. The state of the neural network is reset at each new window for each new prediction.

### 3.3. Tricks for Optimization

Our Deep-Res-Bidir-LSTM for HAR makes it possible to see that the accuracy during testing is much worse than that during training. Overfitting is likely to occur, and balancing the regularization hyper-parameters becomes difficult because they are so numerous. The L2 norm of the weights for weight decay is added in the loss function.

Also, dropout is applied between each layer on the depth axis or, sometimes, just at the output, depending on what is specified in the configuration file, which is another hyper-parameter. Dropout refers to the fact that parts of tensors that are output by the hidden layer are shut down to a zero value to a certain probability for each value in each training epoch, while other values scale up accordingly to keep the same geometric norm of the tensor's values. The inoperative nodes can be regarded as dead nodes (or neurons) that are temporarily not in the network, which means that the weights and biases behind these dead notes temporarily neither learns nor contributes to the predictions during that training step for a batch. The weights are kept intact.

To avoid a sudden leveling off in the accuracy during learning, gradient clipping [22] is added with a maximal gradient step norm of 15. This threshold, $v(v>0)$, for the gradient helps not to overshoot the weight update during training due to having sharp cliffs in the weight space, a characteristic of RNNs;

$$g \leftarrow \frac{gv}{\|g\|}, \quad \text{if } \|g\|>v \tag{6}$$

where $g$ is the gradient and $\|g\|$ is the normed gradient.

Batch normalization [23] can also be useful in training residual connections. The fundamental idea of batch normalization is that layers are simply normalized by mean and variance such that they have a mean of zero and a standard deviation of 1 over the whole batch, so one big rescaling factor multiplies the whole batch, and one big bias is also added. The result is then normalized and offset in a linear fashion. The scaling multiplier $\alpha$ and the offset parameter $\beta$ are learned to rescale inputs in a custom way, and $\beta$ can be initialized to 1, as is commonly done. The formula can be defined as:

$$\hat{x}^{(k)} = \frac{x^{(k)} - E[x^{(k)}]}{\sqrt{Var[x^{(k)}]}},$$
$$y^{(k)} = \alpha^{(k)} \hat{x}^{(k)} + \beta^{(k)} \tag{7}$$

where $x^{(k)}$ means the $k_{th}$ parameter in the parameters vector, and $y^{(k)}$ is the normed value of $x^{(k)}$.

We added many tricks to the network to provide better results. Generally, L2 norm for weight decay and dropout are used to prevent overfitting, and gradient clipping and batch normalization are used to prevent gradient vanishing or explosion as well as overshooting the learning updates.

## 4. Experiments

We tested the Deep-Res-Bidir-LSTM network with the public domain UCI data set [24] and the Opportunity data set [7]. Then, we compared it with the outcomes of other methods and analyzed the results. The computer for testing had an i7 CPU with 8 GB RAM as well as an NVIDIA GTX 960m GPU, which has 640 CUDA cores and 4 GB RAM. The GPU and CPU were used alternatively depending on the size of the neural network, which sometimes exceeded the available amount of memory on the graphics card during training.

### 4.1. Data Sets

The research objects of recognition were activities in daily life. Thus, the benchmark for HAR should meet two conditions: first, it should contain most behavioral classes so it reflects real life. Second, it should abstract features and labels for modeling and calculations. Human actions can be divided into several layers in terms of granularity, such as the gesture layer (including left-arm lift, trunk-back), the action layer (including jumping, running, sitting), and the behavior layer (such as drinking water, typing, sleeping). A good HAR benchmark should include a clear understanding of the hierarchy. There are several open data sets that can be benchmarked, such as the public domain UCI [24], the Opportunity [7], and the KTH data sets [25]. Many studies have used these benchmarks. We chose the public domain UCI and the Opportunity data sets for our experiments. The neural network should be readily adaptable to a new data set with an architecture module and a changeable configuration file that also loads the data set.

**Public domain UCI data set**. Experiments were carried out with a group of 30 volunteers aged 19–48 years. Each person performed six activities (WALKING, WALKING_UPSTAIRS, WALKING_DOWNSTAIRS, SITTING, STANDING, LAYING_DOWN) wearing a smart phone (Samsung Galaxy S II) on the waist. Using its embedded accelerometer and gyroscope, we were able to make three-axial linear acceleration and three-axial angular velocity available at a constant rate of 50 Hz and trimmed into windows of 128 time steps for a 2.56 seconds windows; this was enough to capture two steps, in the case of walking, for the classification. The experiments were video-recorded to label the data manually and obtain balanced classes; the data were of high quality. The data set obtained was partitioned randomly into two sets: 70% of the volunteers were selected for generating the training data, and 30% were selected for generating the test data. Each sample had 561 linear (time-independent) hand-made, preprocessed features from signal analysis (e.g., window's peak frequency), but only nine features were used in our study: triaxial gravity acceleration from the accelerometer (from a 0.3 Hz Butterworth low-pass filter) and triaxial body acceleration and triaxial angular velocity from the gyroscope. These are raw signals with a time component and do not fall in the frequency domain but rather in the time domain. The sensor data were pre-processed by applying denoising median filters, clipping the approximately 20 Hz mark; they were then sampled in fixed-width sliding windows of 2.56 seconds. Those windows were provided with an overlap of 50% to ease training. Additionally, all features were pre-normalized and bounded within [-1, 1].

**Opportunity data set**. The Opportunity data set for HAR from wearable, object, and ambient sensors is a data set devised to benchmark HAR algorithms. The data set includes activities from four subjects; each one has six recorded runs. For each subject, the first five records consist of runs of activities of daily living, characterized by the natural execution of daily activities. The sixth run was a "drill" run, where users executed a scripted sequence of activities. The activities of the user in the scenario were annotated on different levels. Notably, among others, 17 mid-level gesture classes were identified and used for our predictions; this group included the "NULL" class, which is common, for a total of 18 classes. The NULL class rendered the data set highly unbalanced; thus, following previous research [10], we used a weighted F1 score [10]. In total, 242 features from body-worn sensors, object sensors, and ambient sensors were provided for each sample; time stamps in milliseconds, starting from zero and having a sampling rate of 30 Hz, were also provided. Many of those 242 features are not useful for HAR; thus, we used only 113 features, such as DeepConvLSTM [10]. Due to the use of wireless sensors to transfer data, there may be missing data. We used linear interpolation to fill in the missing data. Also, the data were provided with a custom scale and different value ranges and resolutions for each feature; there were sometimes magnitudes of difference according to the cell used. Our architecture used mean and variance (standard deviation) normalization on the z-score scale with a standard deviation of 0.5. Such a small standard deviation is often useful in deep learning [30]. The transition function was defined as follows:

$$x^* = \frac{x - \mu}{\sigma},  \qquad (8)$$

where $\mu$ is mean value and $\sigma$ is the standard deviation. As with DeepConvLSTM, we used a subset from subjects 1 to 3 as a training data set and used the remainder of this subset for the test data, using runs 3 and 4 of subjects 2 and 3 as testing data, for a total of 4 test runs. To obtain comparable results, we did not use the data from subject 4. To summarize, we used only a subset of the full data set to simulate the conditions of the competition, using 113 sensor channels and classifying the 17 categories of output (and the NULL class). Our LSTM's inner representation was always reset to 0 between series. We used mean and variance normalization rather than min-to-max rescaling.

Because of class imbalance in the Opportunity data set, we used the F1 score as a measurement of recognition. The F1 score can be regarded as a weighted average of accuracy and recall; it ranges between 0 and 1. For a dichotomous problem, the F1 score can be defined as follows:

$$F_1 = 2 \frac{prec \times recall}{prec + recall}, \qquad (9)$$

where $prec$ and $recall$ indicate precision and recall, respectively.

However, we needed a multi-class classification in this paper. So, the F1 score was defined as follows:

$$F_1 = 2\sum_c \frac{N_c}{N_{total}} \frac{prec \times recall}{prec + recall}, \tag{8}$$

where $N_c$ is the sample count of class $c$, and $N_{total}$ is the total sample count of the data set.

### 4.2. Hyper-parameters Setting

The hyper-parameters in the Deep-Res-Bidir-LSTM network affect the final result. Generally used methods of tuning parameters include experimental methods, grid searches [26], genetic algorithm (GA) [27], and particle swarm optimization (PSO) [28][29]. As experimental methods involve approximating the value by running many experiments, these methods are time consuming. GA and PSO are heuristic algorithms, and they are limited to dealing with large-scale network. We used grid search, which involved dividing hyper-parameter values into several steps to create a grid of a certain range, then traversing all points of the grid to find the best values for these parameters. We used a grid search, iteratively improved the neural network, and repeated the grid search.

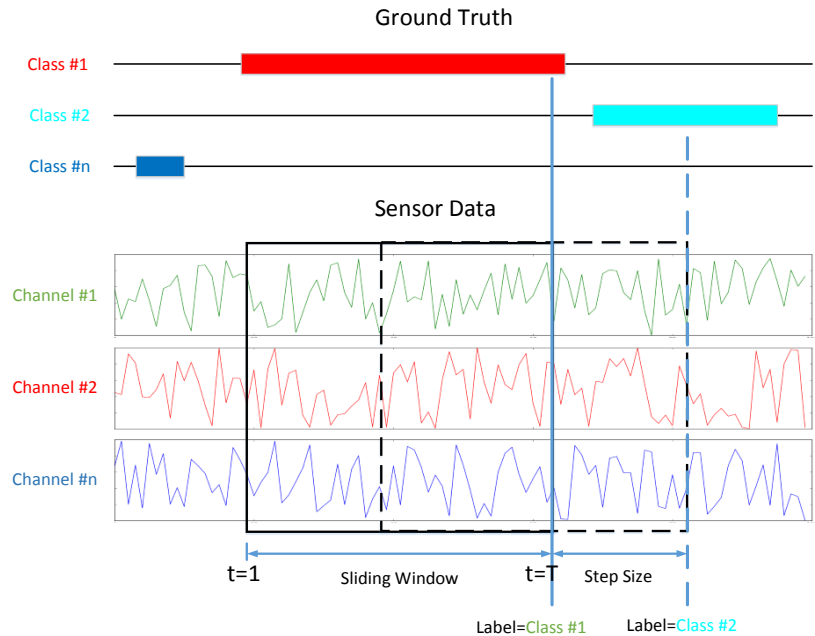

Figure 6. Sliding window. The ground truth represents labels for the classification.

Our LSTM's inner representation was always reset to 0 between series. As shown in Figure 6, for n channels' data, a new sample consisted of a window length series of T. Then, T was moved with a step size to form the next sample, using 50% overlap, which added some redundancy during training and testing. Repeating the operation above yielded a data set suitable for the training. Missing values between labels refer to the Null class. The last moment of the time window was chosen as a label for each sample's classification. Thus, if the class changed throughout the series, only the last time step would account for the classification for training and testing, and the 50% overlap accounted for learning some of the label changes. The label of the last moment was better at reflecting the intent as well as the possibly current behavior. Through a slide window process, the shape of the data set $SequenceNum \times channels$ was converted to $sampleNum \times windowSize \times channels$ with overlap and, therefore, with duplication of the size of the contents.

### 4.3. Results

Table 1. Accuracy and F1 score for each algorithm with the public domain UCI data set. The best result in each column is highlighted in bold.

| Algorithm | Accuracy | F1 score |
|---|---|---|
| MC-SVM [8] | 89.3% | - |
| Baseline LSTM | 90.77% | 90.77% |
| Bidir-LSTM | 91.09% | 91.11% |
| Res-LSTM | 91.55% | 91.51% |
| Deep-Res-Bidir-LSTM | **93.57%** | **93.54%** |

MultiClass Hardware Friendly SVM, or MC-HF-SVM, was proposed by Davide Anguita [8]. It allows better preservation of the life of a smart phone battery than the conventional floating-point-based formulation while maintaining comparable system accuracy levels. The performances of Bidir-LSTM and Res-LSTM were almost the same; both were better than the baseline LSTM, because they are good at information transmission. Bidir-LSTM can get information in both forward and backward passes, and

Res-LSTM uses a highway to transmit information directly. Among the algorithms in Table 1, Deep-Res-Bidir-LSTM achieved the best F1 score, 93.54%, because of both residual connections and bidirectional cells. Comparing accuracy and F1 scores, the two columns are almost the same for each model. We randomly selected a batch while training, and a complete calculation was almost able traverse the entire data set.

Table 2. F1 score with the NULL class of each algorithm with the Opportunity data set

| Algorithm | F1 score |
| --- | --- |
| LDA [7] | 69% |
| QDA [7] | 53% |
| NCC [7] | 51% |
| 1NN [7] | 87% |
| 3NN [7] | 85% |
| UP [7] | 64% |
| NStar [7] | 84% |
| SStar [7] | 86% |
| DBN [9] | 73.0% |
| CNN [9] | 85.1% |
| Deep-Res-Bidir-LSTM | **90.20%** |

It can be seen that the proposed algorithm outperformed the others. It was about 3.68% better in gesture recognition. Due to the dominant Null class, most samples tended to be classified into the Null class. This class imbalance occurred with all algorithms, but its severity, in the F1 score, is seen more for other algorithms.

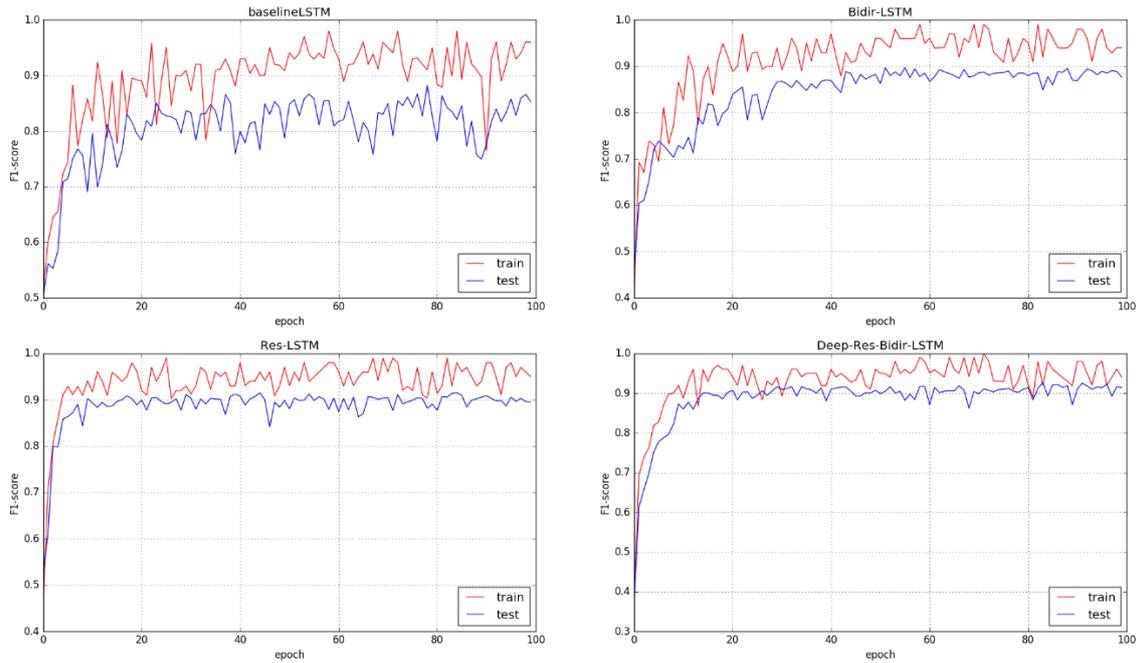

Figure 7. F1 score trends of algorithms. The blue line shows test data, and the red one indicates training data. (a)–(d) show baseline LSTM, Bidir-LSTM, Res-LSTM, and Deep-Res-Bidir-LSTM, respectively.

Figure 7 shows the F1 score trend with the training data and testing data for each model. Generally, when training was finished, both the training and testing results oscillated around a fixed value. Moreover, the results in the four groups were convergent. The amplitude of baseline LSTM was significantly higher than those of the other three. Deep-Res-Bidir-LSTM achieved the best F1 score, ~0.9. Convergence rate can be arranged from slow to fast as baseline LSTM, Bidir-LSTM, Deep-Res-Bidir-LSTM, and Res-LSTM. However, the difference between Deep-Res-Bidir-LSTM and Res-LSTM was very small, and both were obviously different from the others. The results also show that residual connection was outstanding in convergence.

Table 3. Matrix confusion on test data set using Dee-Res-Bidir-LSTM. WK, WU, WD, ST, SD, and LD (representing WALKING, WALKING_UPSTAIRS, WALKING_DOWNSTAIRS, SITTING, STANDING, LAYING_DOWN, respectively).

|     | WL | WU | WD | ST | SD | LD | Recall |
| --- | --- | --- | --- | --- | --- | --- | --- |
| WL | **476** | 1 | 20 | 0 | 0 | 0 | 95.77% |
| WU | 20 | **429** | 21 | 0 | 0 | 0 | 91.28% |
| WD | 14 | 8 | **398** | 0 | 0 | 0 | 94.76% |
| ST | 0 | 5 | 0 | **426** | 33 | 4 | 91.09% |
| SD | 0 | 0 | 0 | 62 | **473** | 0 | 88.41% |
| LD | 0 | 0 | 0 | 0 | 0 | **537** | 100% |
| Precision | 93.33% | 96.84% | 90.66% | 87.30% | 93.48% | 99.26% | 93.57% |

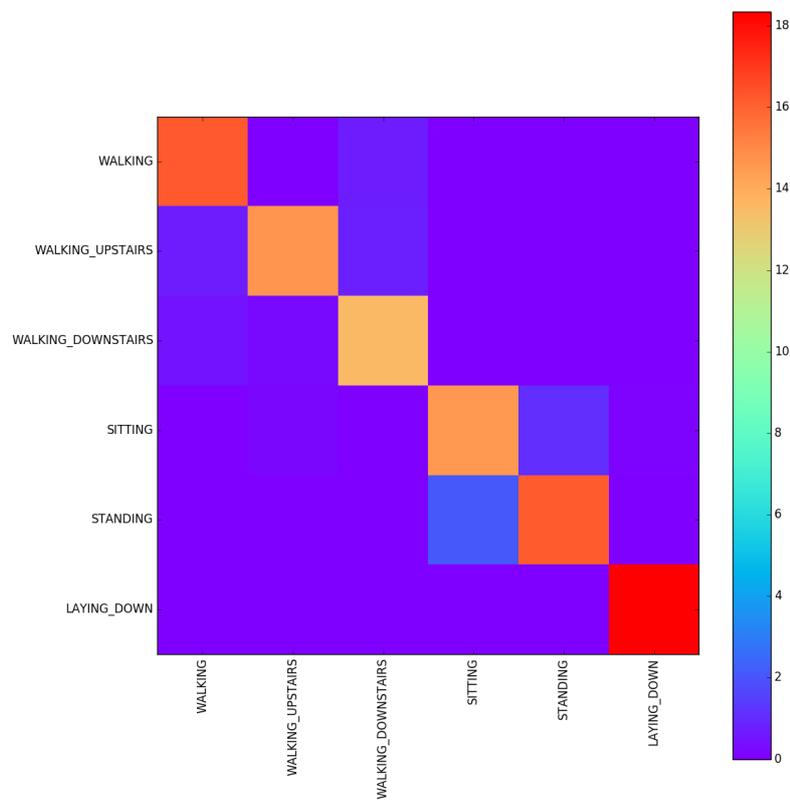

Figure 8. Normalized-to-percent matrix confusion on the test data set using Dee-Res-Bidir-LSTM. The columns represent the predicted classes, and the rows represent the actual classes.

Table 3 shows the confusion matrix of Deep-Res-Bidir-LSTM with testing data. The values of prediction in six classes were in the range of 87.30% to 99.26%, and the values of recall were in the range of 88.41% to 100%. The integral accuracy reached 93.57%. An intuitive confusion matrix is shown in Figure 8. The color from blue to red represents the increasing percentage. It can be seen that the LAYING_DOWN class was recognized best, likely because triaxial acceleration and triaxial angular velocity are quite different from the values in other classes. Standing and sitting were sometimes misrecognized as each other; both involve static behavior. In fact, they are seemingly almost the same from the point of view of a device placed on the belly, which is how the data set was gathered. Similarly, WALKING, WALKING_UPSTAIRS, and WALKING_DOWNSTAIRS all involve dynamic behavior, and there is some confusion among them. However, it is still possible to see a little clustering among these three classes in the matrix.

## 5. Conclusions

In this paper, the significance of HAR research is analyzed, and an overview of emerging methods in the field is provided. LSTM neural networks have been used in many innovations in natural language processing, speech recognition, and weather prediction. This technology was adapted to the HAR task. We proposed the novel framework of the Deep-Res-Bidir-LSTM network. This deep network can enhance learning ability for faster learning in early training. The proposed network also guarantees the validity of information transmission through residual connections (on the depth dimension) and bidirectional cells (on the temporal dimension). In our experiments, the proposed network was able to improve the accuracy, by 4.78%, for the public domain UCI data set and increase the F1 score, by 3.68%, for the Opportunity data set in comparison with previous work.

We also found that window size was a key parameter. Too small a window did not guarantee continuity of information, and too large a window caused classification errors. Usually, 500 ms to 5000 ms will be appropriate for the window size. During model training, the architecture of the network, such as the layers and the cells in each layer, should be determined first, followed by the optimization of hyper-parameters, such as learning rate and the L2 weight decay multiplier. The values of hyper-parameters should be determined according to the specific architecture. For example, 28 cells were sufficient for the public domain UCI data set, but 128 cells were better for the Opportunity data set because it has more features and labels and, thus, increased overall complexity.

Future work should explore a more efficient way to tune parameters. Although the grid search was workable, the searching range must be changed manually, and the values are always fixed. It will be important to find an adaptive way to automatically adjust the searching process and also make the neural network's architecture evolve, such by as automatically reshaping, adding, and removing various layers. Also, exploring the effect of mixing 1D time-based convolutions at one or some points in the LSTM cells might improve results. Finally, applying the Deep-Res-Bidir-LSTM network to other fields could be revealing. A good model should have outstanding generalization. Indeed, focusing on time series prediction problems has value. Problems such as stock prediction and trajectory prediction may be explored.

**Acknowledgments**

We thank Guillaume Chevalier, who supported this work by offering guidance and by contributing to building the neural network architecture. Our Github repository is located at the following address:

https://github.com/guillaume-chevalier/HAR-stacked-residual-bidir-LSTMs.

The code is open source, and we sought to make it readily adaptable to new problems using a new configuration file. We also thank everyone who contributed to building and maintaining the openly available data sets on the UCI Machine Learning Repository; other links [33][34] are cited in the References.

# References

[1] Kalantarian H, Sideris C, Mortazavi B, et al. Dynamic computation offloading for low-power wearable health monitoring systems, IEEE Transactions on Biomedical Engineering. 64(3) (2017) 621–628.

[2] Pantelopoulos A, Bourbakis N G. A survey on wearable sensor-based systems for health monitoring and prognosis, IEEE Transactions on Systems Man & Cybernetics Part C Applications & Reviews. 40(1) (2010) 1–12.

[3] Ahmed S H, Kim D. Named data networking-based smart home, ICT Express. 2(3) (2016) 130–134.

[4] Kumar, Shiu. Ubiquitous smart home system using android application, arXiv preprint, arXiv: 1402.2114, 2014.

[5] Rautaray S, Agrawal A. Vision based hand gesture recognition for human computer interaction: a survey, Artificial Intelligence Review. 43(1) (2015) 1–54.

[6] Yeo H S, Lee B G, Lim H. Hand tracking and gesture recognition system for human–computer interaction using low-cost hardware, Multimedia Tools and Applications. 74(8) (2015) 2687–2715.

[7] Chavarriaga R, Sagha H, Calatroni A, et al. The Opportunity challenge: A benchmark database for on-body sensor-based activity recognition, Pattern Recognition Letters. 34(15) (2013) 2033–2042.

[8] Anguita D, Ghio A, Oneto L, et al. Energy efficient smartphone-based activity recognition using fixed-point arithmetic, Journal of Universal Computerence. 19(9) (2013) 1295–1314.

[9] Yang J B, Nguyen M N, San P P, et al. Deep convolutional neural networks on multichannel time series for human activity recognition, in: International Conference on Artificial Intelligence. AAAI Press, 2015.

[10] Ordóñez F J, Roggen D. Deep Convolutional and LSTM Recurrent Neural Networks for Multimodal Wearable Activity Recognition, Sensors. 16(1) (2016) 115–140.

[11] Wang N, Yeung D Y. Learning a deep compact image representation for visual tracking, in: Advances in Neural Information Processing Systems, 2013.

[12] Dong C, Loy C, He K, et al. Learning a deep convolutional network for image super-resolution, in: European Conference on Computer Vision, 2014.

[13] Graves A, Mohamed A R, Hinton G. Speech recognition with deep recurrent neural networks, in: IEEE International Conference on Acoustics, Speech and Signal Processing. 2013.

[14] Hannun, Awni, et al. Deep speech: Scaling up end-to-end speech recognition, arXiv preprint, arXiv: 1412.5567, 2014.

[15] Abdel-Hamid O, Mohamed A R, Jiang H, et al. Convolutional Neural Networks for Speech Recognition, IEEE/ACM Transactions on Audio Speech & Language Processing. 22(22) (2014) 1533–1545.

[16] Kumar, Ankit, et al. Ask me anything: Dynamic memory networks for natural language processing, in: CoRR, 2015.

[17] Cho, Kyunghyun, et al. Learning phrase representations using RNN encoder-decoder for statistical machine translation, arXiv preprint arXiv: 1406.1078, 2014.

[18] Krizhevsky A, Sutskever I, Hinton G E. ImageNet classification with deep convolutional neural networks, in: International Conference on Neural Information Processing Systems, 2012.

[19] Hochreiter, Sepp, and Jürgen Schmidhuber. Long short-term memory, Neural computation. 9(8) (1997) 1735–1780.

[20] Bengio Y, Simard P, Frasconi P. Learning long-term dependencies with gradient descent is difficult, IEEE Transactions on Neural Networks. 5(2) (1994) 157–166.

[21] He, Kaiming, et al. Deep residual learning for image recognition, in: Proceedings of the IEEE Conference on Computer Vision and Pattern Recognition, 2016.

[22] Pascanu, Razvan, Tomas Mikolov, and Yoshua Bengio. Understanding the exploding gradient problem, in: CoRR, 2012.

[23] Ioffe, Sergey, and Christian Szegedy. Batch normalization: Accelerating deep network training by reducing internal covariate shift, arXiv preprint, arXiv:1502.03167, 2015.

[24] Anguita, Davide, et al. A Public Domain Dataset for Human Activity Recognition using Smartphones, in: ESANN, 2013.

[25] Maclean W J. Spatial Coherence for Visual Motion Analysis, Springer Berlin Heidelberg, 2006.

[26] Yao Y, Zhang L, Liu Y, et al. An improved grid search algorithm and its application in PCA and SVM based face recognition, Journal of Computational Information Systems. 10(3) (2014) 1219–1229.

[27] Levy E, David O E, Netanyahu N S. Netanyahu. Genetic algorithms and deep learning for automatic painter classification, in: proceedings of the 2014 Annual Conference on Genetic and Evolutionary Computation, 2014.

[28] Fornarelli G, Giaquinto A. Adaptive particle swarm optimization for CNN associative memories design, Neurocomputing. 72(16) (2009) 3851–3862.

[29] Syulistyo A R, Purnomo D M J, Rachmadi M F, et al. PARTICLE SWARM OPTIMIZATION (PSO) FOR TRAINING OPTIMIZATION ON CONVOLUTIONAL

NEURAL NETWORK (CNN), Jurnal Ilmu Komputer dan Informasi. 9(1) (2016): 52-58.

[30] Wiesler S, Richard A, Schluter R, et al. Mean-normalized stochastic gradient for large-scale deep learning, in: IEEE International Conference on Acoustics, Speech and Signal Processing, 2014.

[31] Kingma, Diederik, Jimmy B. Adam: A method for stochastic optimization. arXiv preprint, arXiv:1412.6980, 2014.

[32] Wollmer M, Eyben F, Keshet J, et al. Robust discriminative keyword spotting for emotionally colored spontaneous speech using bidirectional LSTM networks, in: IEEE International Conference on Acoustics, Speech and Signal Processing, 2009.

[33] David A, Oneto L. Human activity recognition using smartphone data set, https://archive.ics.uci.edu/ml/datasets/Human+Activity+Recognition+Using+Smartphones, 2012.

[34] Daniel R, Rossi M. Opportunity activity recognition dataset, https://archive.ics.uci.edu/ml/datasets/OPPORTUNITY+Activity+Recognition, 2012.